\documentclass[conference]{IEEEtran}
\IEEEoverridecommandlockouts
\usepackage{cite}
\usepackage{amsmath,amssymb,amsfonts}
\usepackage{algorithmic}
\usepackage{graphicx}
\usepackage{textcomp}
\usepackage{xcolor}
\usepackage{multirow}

\def\BibTeX{{\rm B\kern-.05em{\sc i\kern-.025em b}\kern-.08em
    T\kern-.1667em\lower.7ex\hbox{E}\kern-.125emX}}
\begin{document}

\title{Accuracy Improvement of Object Detection in VVC Coded Video Using YOLO-v7 Features\\
\thanks{These research results were obtained from the commissioned
research (No.05101) by National Institute of Information and
Communications Technology (NICT), Japan.}}

\makeatletter
\newcommand{\linebreakand}{%
  \end{@IEEEauthorhalign}
  \hfill\mbox{}\par
  \mbox{}\hfill\begin{@IEEEauthorhalign}
}
\makeatother

\author{\IEEEauthorblockN{1\textsuperscript{st} Takahiro Shindo}
\IEEEauthorblockA{\textit{School of Fundamental Science and Engineering,} \\
\textit{Waseda University}\\
Tokyo, Japan \\
taka\_s0265@ruri.waseda.jp}
\and
\IEEEauthorblockN{2\textsuperscript{nd} Taiju Watanabe}
\IEEEauthorblockA{\textit{School of Fundamental Science and Engineering, } \\
\textit{Waseda University}\\
Tokyo, Japan \\
lvpurin@fuji.waseda.jp}
\linebreakand 
\IEEEauthorblockN{3\textsuperscript{rd} Kein Yamada}
\IEEEauthorblockA{\textit{School of Fundamental Science and Engineering, } \\
\textit{Waseda University}\\
Tokyo, Japan \\
stslm738.ymd@toki.waseda.jp}
\and 
\IEEEauthorblockN{4\textsuperscript{th} Hiroshi Watanabe}
\IEEEauthorblockA{\textit{Graduate School of Fundamental Science and Engineering, } \\
\textit{Waseda University}\\
Tokyo, Japan \\
hiroshi.watanabe@waseda.jp}
}

\maketitle

\begin{abstract}
With advances in image recognition technology based on deep learning, automatic video analysis by Artificial Intelligence is becoming more widespread.
As the amount of video used for image recognition increases, efficient compression methods for such video data are necessary.
In general, when the image quality deteriorates due to image encoding, the image recognition accuracy also falls.
Therefore, in this paper, we propose a neural-network-based approach to improve image recognition accuracy, especially the object detection accuracy by applying post-processing to the encoded video. 
Versatile Video Coding (VVC) will be used for the video compression method, since it is the latest video coding method with the best encoding performance.
The neural network is trained using the features of YOLO-v7, the latest object detection model. 
By using VVC as the video coding method and YOLO-v7 as the detection model, high object detection accuracy is achieved even at low bit rates. 
Experimental results show that the combination of the proposed method and VVC achieves better coding performance than regular VVC in object detection accuracy.
\end{abstract}


\begin{figure*}[bt]
    \centerline{\includegraphics[width=2\columnwidth]{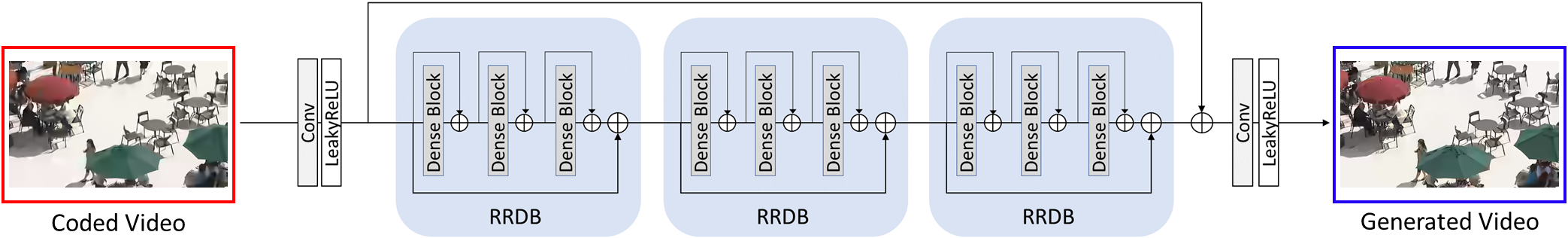}}
    \caption{Model strcture of the post-processing for VVC encoded video.}
    \label{fig:structure}
    \end{figure*}

\section{Introduction}
In recent years, the amount of videos used for image recognition has been rapidly increasing.
Most of the videos captured by consumer video cameras are encoded and used for human vision and image recognition.
Thus, new video coding method and its standardization for this purpose are desired. 
The Moving Picture Experts Group (MPEG) named this research area as ``Video Coding for Machines (VCM)''.
MPEG is attempting to standardize VCM from two perspectives: feature coding and video coding. 
Among them, there are two main approaches for standardization based on video coding.
The first one is to compress video for image recognition.
The information of video required for image recognition is less than that required to create video for viewing \cite{b3}. 
Therefore, higher compression ratio can be achieved when encoding scheme is designed only for image recognition. 
However, the application of video compressed for image recognition is limited because it does not contain enough information necessary to reconstruct the video for viewing. 
The second approach is to compress video for viewing and then converting it for machines. 
For the video compression methods for viewing, High Efficiency Video Coding (HEVC) \cite{b5} and VVC \cite{b6} may be utilized.
After using these video coding methods, post-processing is performed using neural networks. 
This approach cannot go beyond the compression ratio of existing coding methods such as VVC.
However, video coding for both viewing and image recognition can be achieved with a single video encoding. 
This paper seeks the way to improve the accuracy of image recognition in VVC coded video by performing post-processing.

Recently, the technology of image recognition has grown remarkably.
In particular, the accuracy of object detection has improved dramatically. 
However, a method of VCM using the latest object detection model has not yet been considered.
Therefore, we propose a method of processing encoded video for latest object detection models.
The model used for object detection is YOLO-v7 \cite{b8}.
YOLO \cite{b7} is one of the most popular object detection models, and YOLO-v7 combines high detection accuracy with fast detection speed.
The model used for video coding is VVC, which is the latest video coding standard.
We process the VVC coded video using Convolutional Neural Network (CNN) to improve the object detection accuracy.
The proposed neural network to process the coded video is shown in Fig. \ref{fig:structure}.
In VCM, it is necessary to understand the characteristics of neural networks that perform image recognition tasks and perform corresponding video compression and processing. 
In our method, the features of YOLO-v7 are extracted to train this neural network used for processing the encoded video. 
This method enables to create videos which preserve the information necessary for object detection by YOLO-v7. 
Experiments show that the proposed video processing can improve the accuracy of object detection in the encoded video.

The rest of this paper is organized as follows: Section 2 and Section 3 describe related work and the proposed method, respectively. 
Section 4 discusses the experiments and results, and the last section presents the conclusion.

\section{Related Works}

\subsection{Video Coding for Human Vision}
Video compression technology is necessary for sending and receiving video within limited communication resources. 
Research and development of technologies to encode video for human vision has been conducted for a long time, and many video coding standards have been created. 
Among them, VVC is the latest video coding standard, the first version of which was completed in July 2020 \cite{b6}. 
Similar to previous coding methods, it combines intraframe and interframe prediction and operates based on hand-crafted algorithms.

On the other hand, research on video coding using neural networks has also been active in recent years.
For example, there are video compression models using RNN \cite{b9}, image compression models using GAN \cite{b10}, and interframe prediction models \cite{b11,b12}. 
CNN-based video coding methods for both intraframe and interframe prediction are also proposed \cite{b13}. Furthermore, the latest models show comparable coding performance to that of VVC \cite{b14}. 

Many studies have been conducted to improve the quality of the coded video by post-processing \cite{b15,b16,b32,b33} using neural networks. 
In VVC and HEVC, block-based motion compensation and transformation is performed.
This method causes block noise in coded images, which deteriorate the quality of these images. 
To reduce the block noise, a deblocking filter is employed.
However, removing the noise is not always possible. 
Therefore, a model for removing coding noise using neural networks has been studied.

\subsection{Video Coding for Machines}
As the accuracy of image recognition improves, there are increasing opportunities for machines to perform video analysis \cite{b17}. 
For this reason, coding techniques for image recognition are attracting attention, and the movement towards standardization is accelerating. 
The information in video required for machines is considered to be different from the one for humans to view images \cite{b3}. 
Considering the difference, it is important to capture the characteristics of the information in video, which is necessary for image recognition.
Therefore, video information should be extracted depending on the purpose of the video.

Many studies have been conducted to extract the video information that is necessary for object detection, which is one of the popular image recognition tasks. 
One video coding method extracts the information necessary for YOLO9000 \cite{b18}.
In this model, the video is input to YOLO9000 before coding, and the obtained features are used when the video is encoded \cite{b19}. 
The features represent the part of the image that YOLO9000 places an attention on. 
The encoding model allocates more bits to this part of the image to decode images that are useful for object detection.
The other method integrates video coding model and R-CNN \cite{b20}.
In this coding model, CNN is trained using the detection results of R-CNN \cite{b21}. 
Both have higher object detection accuracy and higher video compression ratio than existing video coding methods for human vision.
Models that compress video information for multi-task are also emerging. 
MSFC \cite{b22} is an image coding model for object detection and segmentation.
The neural network used for image compression is trained using losses computed from object detection and segmentation results. 
Although it is not as accurate as JPEG \cite{b34} in terms of image recognition accuracy, it significantly outperforms JPEG in terms of encoding efficiency.

In addition, research is ongoing to apply video coding methods for human vision, such as VVC, to coding methods for image recognition. 
In some experiments, VVC is used to encode features extracted from videos using neural networks, and these experiments show that VVC can be used effectively in VCM as well.

\begin{figure*}[hbt]
    \centerline{\includegraphics[width=2\columnwidth]{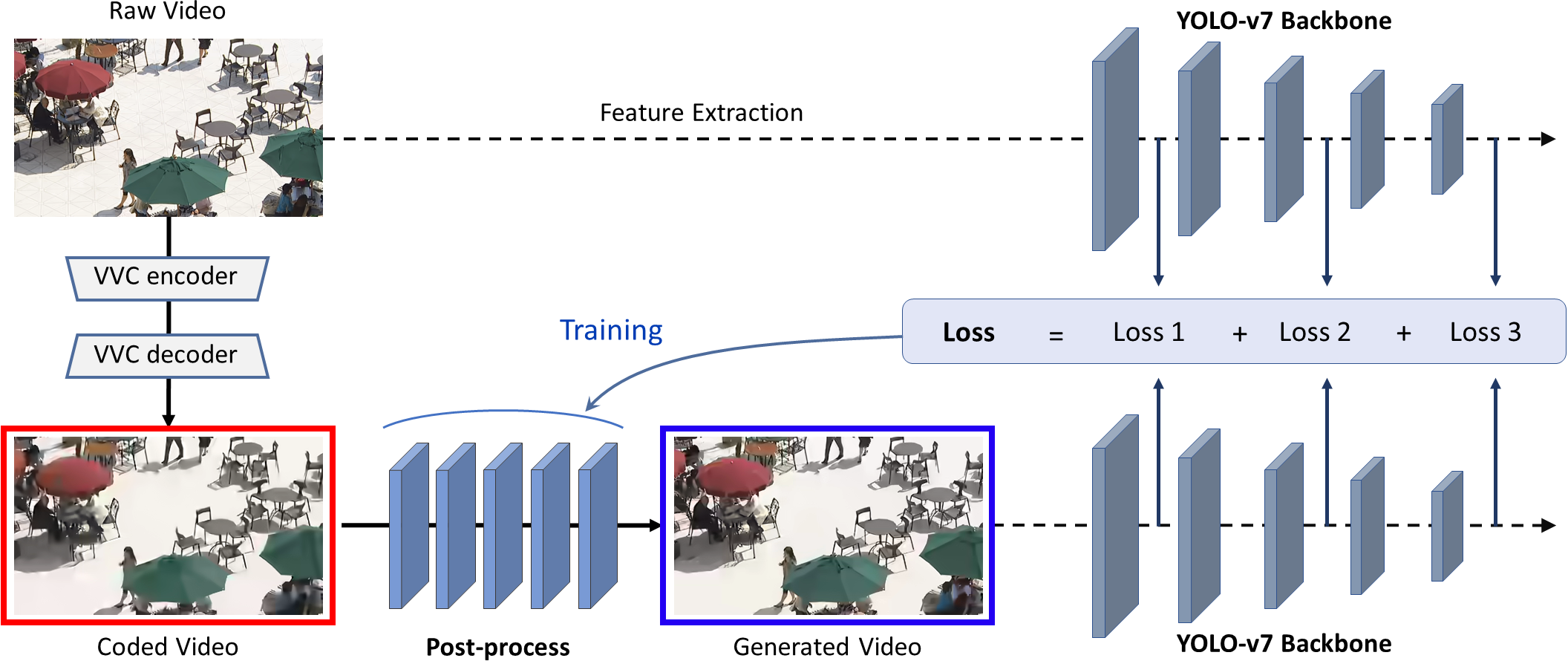}}
    \caption{Training process of proposed neural network.}
    \label{fig:process}
    \end{figure*}

\section{Proposed Method}
\subsection{Approach}
Our approach is to improve the accuracy of object detection by processing the encoded video with a neural network.
The encoded video for human vision is converted to the video for object detection by post-processing.
VVC is used as the video coding method, and YOLO-v7 is utilized as the object detection model. 
The proposed video processing method is based on neural network, and its structure follows the generator of ESRGAN \cite{b23}. 
The proposed model is trained using video features extracted from pre-trained YOLO-v7. 
Processing VVC coded video using the proposed method can achieve high object detection accuracy at low bitrates.
    
\subsection{Model Structure}
Some neural network structures for post-processing of encoded video are based on ResNet \cite{b24}, and some are based on U-Net \cite{b25}. 
These model structures are versatile and have been employed in many models that perform image recognition tasks.
ResNet is a model based on residual blocks and is used in the generator of SRGAN \cite{b26}. 
The network proposed in this paper is based on the residual in residual dense block (RRDB) used in the generator of ESRGAN. 
Same as SRGAN, ESRGAN is a model for image super-resolution. 
The proposed model of the post-processing is shown in Fig. \ref{fig:structure}.
The use of RRDB allows to reproduce more detailed patterns in the generated image by neural network than that of residual block. 
The characteristic of RRDB is also useful in processing of encoded video. 
This is because it helps reconstruct the details of the image.
Our model consists of three RRDBs, two convolutional layers, and two activation functions.

\subsection{Loss Function}
Results of image recognition tasks are generally used to train neural networks which compress the video for image recognition. 
For example, in the studies by S. Wang et al. \cite{b21} and Z. Zhang et al. \cite{b22} the results of object detection by R-CNN are used to train the image compression models to improve the accuracy of object detection by R-CNN. 
In our study, YOLO-v7, the latest version of YOLO, is used as the object detection model. 
In order to improve the object detection accuracy of YOLO-v7 in encoded videos, it is effective to process these videos using the features of the pre-trained YOLO-v7.
We process the encoded videos with a CNN-based neural network trained with YOLO-v7 features. 
In this training, we extract the three kind of features of YOLO-v7 from the backbone of the trained model.
For the loss calculation, we use the mean squared error (MSE) between the features of raw video and that of output video of our proposed model. 
The training process of the proposed model is shown in Fig. \ref{fig:process}.
The loss function used for training is extracted as
\begin{eqnarray}
  \label{eq:loss}
  Loss &=& MSE(yolo(I_{raw}), yolo(I_{output})),
\end{eqnarray}
where $yolo$ indicates the feature extractor of YOLO-v7.
$I_{raw}$ and $I_{output}$ indicate the raw video and the output video.

\begin{figure*}[t]
    \centerline{\includegraphics[width=2\columnwidth]{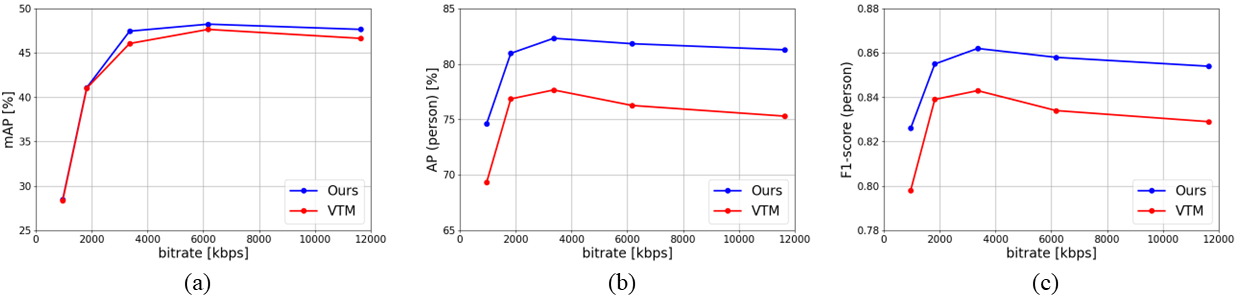}}
    \caption{Evaluation result of PeopleOnStreet sequence in terms of (a) rate-mAP; (b) rate-AP(person); (c) rate-F1(person).}
    \label{fig:people}
    \end{figure*}

\begin{figure*}[t]
    \centerline{\includegraphics[width=2\columnwidth]{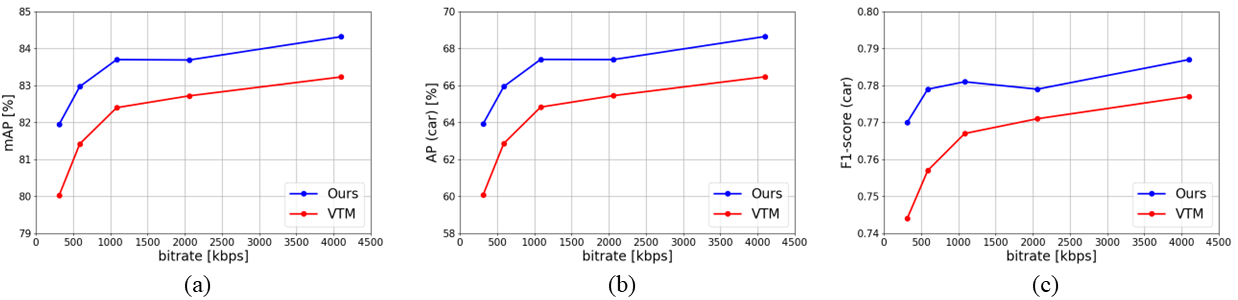}}
    \caption{Evaluation result of Traffic sequence in terms of (a) rate-mAP; (b) rate-AP(car); (c) rate-F1(car).}
    \label{fig:traffic}
    \end{figure*}

\begin{figure}[bt]
    \centerline{\includegraphics[width=1\columnwidth]{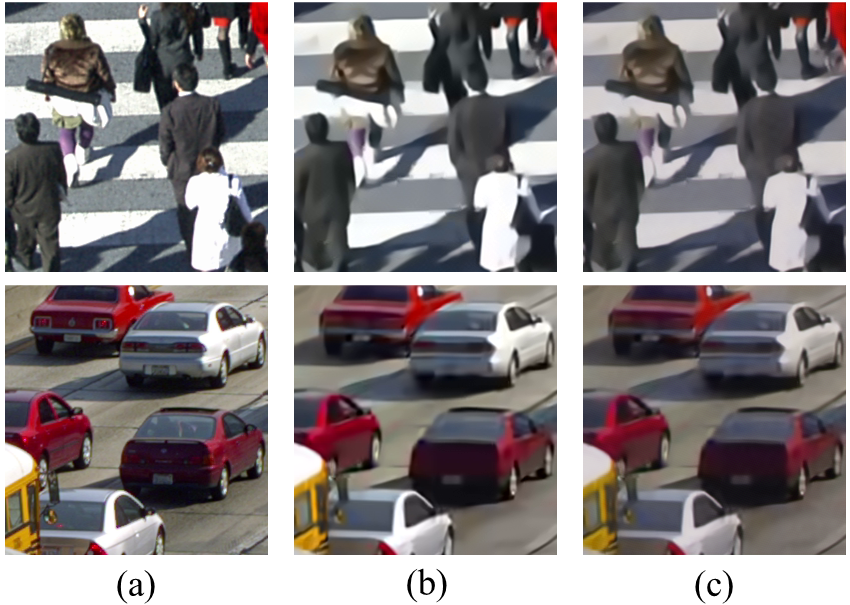}}
    \caption{Comparison of frame images (240x270). (a) raw frame; (b) input frame encoded with VVC (QP:47); (c) output frame of the proposed method. The upper images are part of PeopleOnStreet sequence. The lower images are part of Traffic secuence.}
    \label{fig:frame}
    \end{figure}

\section{Experiments}
\subsection{Training Details}

In this section, we present a training process of proposed neural network for processing the encoded video. 
The datasets used for training are SJTU \cite{b27}, UVG \cite{b28}, and MCML-4K-UHD \cite{b29}. 
All these datasets contain some 4K (3840x2160) raw video sequences. 
We select 30 sequences from these datasets and encode these by using VTM10.0 \cite{b30}. 
The configration of the frame reference method is ``random access'', and the quantization parameters (QP) are 27, 32, 37, 42, and 47. 
These 150 types of encoded video created in this way are used as input video for the proposed neural network. 
The loss function is as shown in (\ref{eq:loss}), and features obtained from the trained model of YOLO-v7 are used. 
The optimization function is Adam and the learning rate is 1e-5.
GPU used for training neural network is GeForce GTX 1080ti. 

\subsection{Evaluation Method}
We perform object detection using YOLO-v7 on the VVC encoded video and the output video of the proposed neural network.
In order to measure the object detection accuracy, we need video sequences with object annotations. 
Therefore, we use SFU-HW-Objects-v1 dataset \cite{b31} for evaluation.
This dataset is one of the few datasets in which object annotations are assigned to raw video. 
It contains object annotations for 18 raw video sequences. 
These sequences are classified into five classes according to the image size and the characteristics of the video. 
Since the image size of the video sequences used in training is 4K, we use sequences of Class A, B and C, which have a larger image size compared to other classes. 
The details of our test sequences are shown in Table \ref{tab:seq}.
These sequenses are also encoded using VTM 10.0. 
The configration of the frame reference method is ``random access''.
Two sequences of Class A are encoded with five QP (27, 32, 37, 42, 47), and other sequences are encoded with three QP (37, 42, 47). 
We adopt the proposed post-processing method to the encoded video, and the object detection accuracy before and after the proposed processing is compared. 
The object detection model is pre-trained YOLO-v7, and the confidence threshold is set to 0.25. Average Precision (AP) and F1-score are used as evaluation metrics.
When caluculating the value of AP, Intersection over Union (IoU) threshold is always set to 0.5.

\begin{table}[t]
  \centering
  \caption{List of test sequences.} \label{tab:seq}
  \begin{tabular}{c|c|c|c}
    \hline
    class & sequence name & size & frame number\\
    \hline
    \hline
    A & PeopleOnStreet & 2560x1600 & 150\\
    A & Traffic & 2560x1600 & 150\\
    B & BQTerrace & 1920x1080 & 600\\
    B & BasketballDrive & 1920x1080 & 500\\
    B & ParkScene & 1920x1080 & 240\\
    C & BQMall & 832x480 & 600\\
    C & BasketballDrill & 832x480 & 500\\
    C & PartyScene & 832x480 & 500\\
    C & RaceHorsesC & 832x480 & 300\\
    \hline
  \end{tabular}
\end{table}

\subsection{Results}
The object detection results in PeopleOnStreet sequence are shown in Fig. \ref{fig:people}. 
Since ``person'' accounted for 97\% of the objects in PeopleOnStreet sequence, the detection results of person is also shown in Fig. \ref{fig:people}. 
All three graphs show the relationship between detection accuracy and bitrate. 
Graph (a) represents mean Average Precision (mAP), graph (b) AP of the person detection, and graph (c) F1-score of the person detection. 
In this sequence, the proposed method slightly increased the value of mAP, and significantly increased the accuracy of person detection. 
The value of AP of person detection was improved by about 5 percentage points regardless of QP, and F1-score was improved by about 0.02 to 0.03.
The transformation of the frame image of PeopleOnStreet sequence by the proposed method are shown in Fig. \ref{fig:frame}.
Image (b) is the frame image of VVC encoded video, and (c) is that of the output video of the proposed neural network.
This figure shows that the proposed post-processing slightly changes the color tone of the frame image.

The object detection results in Traffic sequence are shown in Fig. \ref{fig:traffic}. 
Since the percentage of objects occupied by ``car'' in Traffic sequence is 99\%, the detection results of ``car'' is also shown in Fig. \ref{fig:traffic}. 
All these graphs also show the relationship between detection accuracy and bitrate.
Graph (a) represents the value of mAP.
Graph (b) and (c) represent the value of AP and F1-score of the car detection, respectively. 
In Traffic sequence, the value of mAP was improved by proposed method, and AP of car detection was improved by 2 to 3 percentage points and F1-score by about 0.01 to 0.02.
The transformation of the frame image of Traffic sequence by the proposed method are shown in Fig. \ref{fig:frame}.
In the case of Traffic sequence, as in the case of PeopleOnStreet sequence, we can confirm the change in the color tone of the frame image due to the proposed image processing.

\begin{figure}[t]
    \centerline{\includegraphics[width=1\columnwidth]{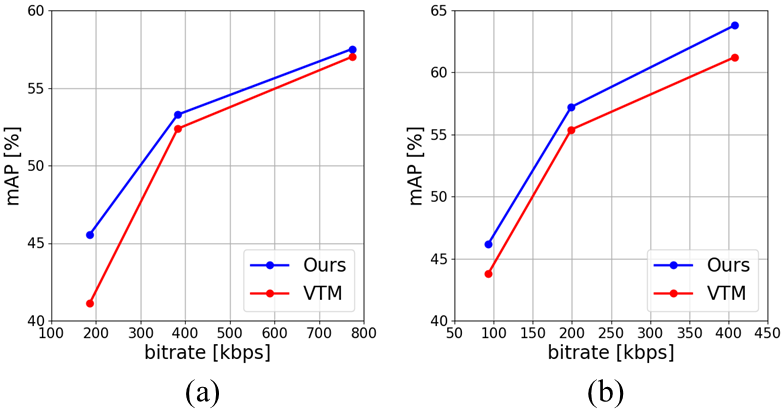}}
    \caption{(a) Evaluation result of class B sequence in terms of rate-mAP; (b) Evaluation result of class C sequence in terms of rate-mAP.}
    \label{fig:classBC}
    \end{figure}

Furthermore, the object detection accuracy in three sequences in Class B and four sequences in Class C are shown in Fig. \ref{fig:classBC}. 
The graph (a) shows the relationship between mAP and bitrate for three sequences of Class B, and the graph (b) shows the relationship between mAP and bitrate for four sequences of Class C. 
In both cases, the object detection accuracy is improved for all QP.

Table \ref{tab:res} summarizes these results. 
For the sequences listed in Table \ref{tab:seq}, the object detection accuracy (mAP) when encoded with VVC and when the proposed post-processing is performed on them are shown. 
The degree of improvement in mAP values are also shown in this table as a gap. 
For the two sequences of class A, the values of mAP are not so improved by the proposed method compared to the AP values of person and car.
The shortage of the training dataset is one cause of this results.
The sequences of training datasets include many persons and cars.
However, some objects, such as umbrella, sport ball and chair, are not included, even though it is included in test sequences.
Nevertheless, Table \ref{tab:res} shows that mAP improved for all QP of all classes. 
The reason for this improvement of mAP value is the enhancement in AP values of person, car, and some other objects.
From these results, our proposed method is effective for improving the object detection accuracy in VVC coded video.

\begin{table}[t]
    \centering
    \caption{Results of mAP values.} \label{tab:res}
    \begin{tabular}{c|c|ccccc}
        \hline
        \multirow{2}{*}{class} & \multirow{2}{*}{method} & \multicolumn{5}{c}{QP} \\
                                &  & 27 & 32 & 37 & 42 & 47 \\
        \hline
        \hline
        \multirow{3}{*}{A} & VTM        & 64.94    & 65.19    & 64.23    & 61.23    & 54.19\\
                            & Ours       & 65.99    & 65.97    & 65.58    & 62.08    & 55.20\\
                            & gap        & +1.05    & +0.78    & +1.35    & +0.85    & +1.01\\
        \hline
        \multirow{3}{*}{B} & VTM        & -        & -        & 57.01    & 52.37    & 41.12\\
                            & Ours       & -        & -        & 57.51    & 53.28    & 45.55\\
                            & gap        & -        & -        & +0.50    & +0.91    & +4.43\\
        \hline
        \multirow{3}{*}{C} & VTM        & -        & -        & 61.23    & 55.39    & 43.77\\
                            & Ours       & -        & -        & 63.80    & 57.12    & 46.16\\
                            & gap        & -        & -        & +2.57    & +1.82    & +2.39\\
        \hline
    \end{tabular}
    \end{table}

\section{Conclusion}
We propose a method to improve the accuracy of object detection by YOLO-v7 in VVC coded video.
In the experiments, we show that the object detection accuracy in VVC coded video can be improved by proposed post-processing using YOLO-v7 features. 
By applying the proposed method to a video coded for human vision, the video can be converted to a video for YOLO-v7.
The converted video has a different color tone from the original encoded video, which facilitates object detection by YOLO-v7.
Furthermore, depending on how the method is used, the video can be adapted to its purpose, either for human vision or for object detection.
In both cases, high compression ratio can be achieved using VVC as the video compression method, and high object detection accuracy is achieved by using pre-trained YOLO-v7 as the object detection model.
For future works, more effective methods of extracting video information required for image recognition will be explored for further enhancement.

\vspace{12pt}

\end{document}